%% file: 00_MAIN_rsc-articletemplate.tex
\documentclass[twoside,twocolumn,9pt]{article}
\usepackage{extsizes}
\usepackage[super,sort&compress,comma]{natbib} 
\usepackage[version=3]{mhchem}
\usepackage[left=1.5cm, right=1.5cm, top=1.785cm, bottom=2.0cm]{geometry}
\usepackage{balance}
\usepackage{mathptmx}
\usepackage{sectsty}
\usepackage{graphicx} 
\usepackage{lastpage}
\usepackage[format=plain,justification=justified,singlelinecheck=false,font={stretch=1.125,small,sf},labelfont=bf,labelsep=space]{caption}
\usepackage{float}
\usepackage{fancyhdr}
\usepackage{fnpos}
\usepackage{microtype}
\usepackage{inconsolata}
\usepackage[english]{babel}
\addto{\captionsenglish}{%
  
}
\usepackage{array}
\usepackage{droidsans}
\usepackage{charter}
\usepackage[T1]{fontenc}
\usepackage[usenames,dvipsnames]{xcolor}
\usepackage{setspace}
\usepackage[compact]{titlesec}
\usepackage{hyperref}

\usepackage{epstopdf}

\definecolor{cream}{RGB}{222,217,201}

\input{_custom-things}

\begin{document}

\input{_style-addons}

\section{Introduction}
\label{sec:introduction}
Transformer-based deep learning techniques have revolutionised the field of Natural Language Processing (NLP) in recent years. They are increasingly applied to chemical sciences where sequence representations of molecular structure, such as SMILES and SELFIES\cite{selfies} bear similarity to language sequences. This makes it possible to adopt NLP algorithms to proces and analyse molecules in a similar fashion as they are used to process and analyse text. This approach can be used for a wide range of tasks, such as molecular property prediction and data-driven molecular structure generation.

The original Transformer architecture, introduced in the seminal paper of Vaswani et al,\cite{Vaswani:2017attention} contains two main components: \textbf{1)} the encoder and \textbf{2)} the decoder. All subsequent NLP models share some relationship with these components. For example, the widely used BERT (Bidirectional Encoder Representations from Transformers) only has the encoder component, while OpenAI’s GPT models and other very recent Large Language Models (LLMs) such as the Llama family~\cite{llama} or Gemini~\cite{gemini} only have the decoder component. The encoder-decoder framework is ideally suited for Sequence-to-Sequence (Seq2Seq) learning, often referred to as text-to-text processing in NLP. In this framework, the encoder component captures the context of input sequences and sends it to the decoder which then generates output sequences.

Several studies have adapted seq2seq models for chemical reaction prediction tasks. Notably, the pioneering Molecular Transformer model by Schwaller et al\cite{Schwaller:2019molecular} and the recent T5Chem model \cite{Lu:2022unified} have achieved impressive performance. T5Chem adapted Google’s T5 (``Text-To-Text Transfer Transformer'') NLP model\cite{t5} to chemical data represented in the SMILES format. T5 closely aligns with the encoder-decoder structure used in the original Transformer model while introduces the ``Text-to-Text'' framework which feeds text sequence (in a natural language such as English) as input and then generates text as output. This allows the same model to handle a variety of tasks simultaneously. To perform a specific task, a task-specific prefix is added to the input sequence, tailoring the model's output. T5Chem pretrained the T5 encoder-decoder architecture with 97M SMILES from PubChem molecules and the \texttt{USPTO\_500\_MT} dataset, creating a multi-task reaction prediction model for five different types of reaction tasks. For example, it uses the task-specific prefix \textit{``Product:''} for reaction product prediction, and a prefix \textit{``Reactants:''} for single-step retrosynthesis. The advantage of multi-task learning is that it allows for simultaneous learning of multiple tasks by leveraging similarities between tasks and offer improved data efficiency, and fast learning without the need to predetermine a single specific prediction task.


T5Chem and a few other similar models \cite{sagawa2023reactiont5, doi:10.1021/acs.jcim.4c00292, doi:10.1021/acs.chemmater.3c01406} have demonstrated the feasibility of a seq2seq framework for a variety of predictions in organic chemistry. However, several crucial issues have not been explored to enable more effective and accurate models. In the present study, we have trained multiple variants of two state-of-the-art seq2seq language models, namely instruction-tuned models Flan-T5\cite{flant5-1} and tokenisation-free byte-level models ByT5\cite{byt5} for the standard organic reaction prediction tasks. With these two model architectures, our aim is to conduct a systematic empirical study on the following aspects:

\rparagraphnodot{1) Adapted and adequate input preprocessing and tokenisation} that reaches beyond natural language towards molecular structure. Tokenisation is usually the first step to train an NLP model. It is the process of breaking a sequence into discrete elements, called ‘tokens’ which are then converted to vectors/embeddings for machine learning models. In NLP, most pretrained language models over the past few years rely on tokenisation performed at sub-word level, as it is effective with frequent tokens, capable of grouping sub-words while having some ability to deal with unknown words. However, sub-word tokenisation is still limited at dealing with variants in spelling (e.g. typos) and unknown characters (e.g. from other languages). Recent approaches, such as ByT5, a variant of the multilingual T5 model which disposes of subword-level tokenisation, have shown the viability of token-free models which were trained on characters in the form of their UTF-8 byte encodings. ByT5 uses a standard Transformer architecture but is `tokenisation-free' as it does not rely on a learned vocabulary to map words or sub-word units to tokens.


In addition, it has been demonstrated in NLP research that ByT5 is significantly more robust to noise in the data and performs better on tasks that are sensitive to spelling, grammar errors and ambiguous expressions, such as text on social media platforms. In terms of data noise, chemical reaction dataset may have a similarity with text on social media. For example, USPTO (United States Patent Office), the largest open-source chemical reaction dataset, contains noise: in this context, data noise is defined as incomplete reaction entry with missing or incorrect reactants, reagents and products and could be quite common in all chemical reaction dataset due to the nature of how chemists record reactions, i.e. focusing on only the main product, and leaving unvaried reagents out when recording many similar reactions.\cite{toniato2021unassisted} Our work will thus implicitly assess whether ByT5’s advantage over dealing with noise in NLP data can be translated to better handling of noise in the chemical reaction data.

\rparagraphnodot{2) Training data efficiency}, i.e., how much annotated data is required in fine-tuning to generalise well to new sequences when working with chemical reaction datasets using seq2seq models.

\rparagraphnodot{3) The use and impact of pretraining}. T5-style language models are pretrained only on language data and/or language tasks, therefore are not ``SMILES-aware''. T5Chem relies on self-supervised pretraining using 97M SMILES to learn the chemical space, which can be extremely GPU-intensive. Recently, language models that pretrained on both language and chemical data have emerged. This cross-domain approach, adopted by models such as MolT5\cite{molt5} and nach0\cite{nach0}, creates a shared representation space. It would be interesting to assess whether such hybrid pretraining offers better initialisation points for task-specific fine-tuning for reaction prediction tasks.

\rparagraphnodot{4) Various other important modelling aspects} that can impact the final task performance, e.g., model size, and decoding algorithm at inference. 

\section{Methodology}
\label{sec:methodology}
\begin{figure*}
    \centering
    \includegraphics[width=1\linewidth]{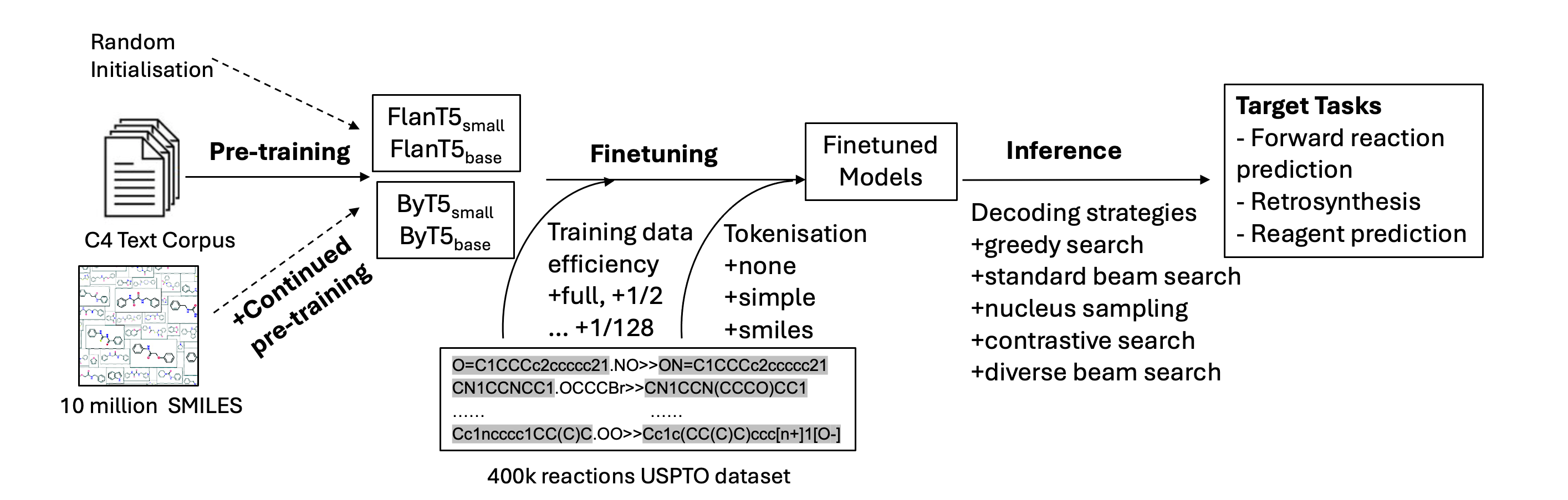}
    \caption{Illustration of the key areas explored along the flow of pretraining, fine-tuning and inference in our work}
    \label{fig:workflow}
\end{figure*}

In this work, we aim to answer a range of questions related to the now established ability \cite{bartsmiles,molt5,Lu:2022unified,nach0} to apply encoder-decoder neural architectures, originally devised for language tasks in NLP research \cite{t5}, to solve tasks related to organic reaction prediction. We vary models, modelling choices, as well as training and evaluation setups across several dimensions of comparison, which we outline next, and also summarise in Figure~\ref{fig:workflow}.

\rparagraph{Model Architectures}
All the models follow the standard Transformer-based encoder-decoder structure where the input sequence is fed to the model (e.g., a sentence in a natural language in case of language models or SMILES for organic reaction prediction), based on the original T5 architecture~\cite{t5}. 

One of the central questions we aim to answer in this work is: can encoder-decoder models, originally pretrained only on language data and/or a variety of language tasks, be effectively specialised to organic reaction prediction tasks via task-specific fine-tuning? Will the model learn to encode and generate SMILES although it was originally pretrained for encoding and generating natural language?  To this end, our starting points are different flavours of the T5-style, all pretrained on language data:
\begin{itemize}[leftmargin=*]
\item The original \textbf{T5} model \cite{t5}, pretrained on the CommonCrawl-based C4 corpus covering $\sim$356B word tokens, via the span-mask denoising objective;
\item The \textbf{FlanT5} model \cite{flant5-1,flant5-2} is an instruction-tuned language model that starts from the pretrained T5 model of the same size, and then `instruction-tunes' it on supervised data of 1,800+ NLP tasks (Flan stands for \textbf{F}inetuning \textbf{Lan}guage models).\footnote{Instruction-tuning is a specialised form of fine-tuning in which a model is fine-tuned using pairs of input-output instructions, enable it to learn specific tasks guided by these instructions.} It typically exhibits better performance than the underlying T5 model across a range of NLP tasks. It can also be used in a standard text-to-text fashion with task-specific fine-tuning if an `empty' instruction is provided to the model (i.e., only the input sequence without an additional task description is provided). This is how we use the model for single task-specific fine-tuning.
\item The \textbf{ByT5} model \cite{byt5} obtains the same architecture, but disposes of standard subword-level tokenisation (see the next paragraph) and processes text as sequences of raw (UTF-8) bytes. Being originally designed to enhance multilingual NLP models, it was pretrained on the multilingual mC4 corpus spanning 101 diverse languages, again relying on the span-mask denoising objective.
\end{itemize}

All the models come in different sizes (in terms of model parameters), and due to high computational demands we mostly focus on benchmarking their \textit{Small} and \textit{Base} variants. The \textit{Small} variant of T5 and FlanT5 comprises $\sim$60M parameters, while \textit{Base} covers 220M parameters. ByT5 variants with the same label are not directly comparable to T5/FlanT5 as they contain a larger number of parameters: \textit{Small} is 300M, and \textit{Base} is 582M parameters.

Assuming the existence of task-specific training data for organic reaction prediction tasks, we also evaluate whether language-specific (and thus `chemical domain-incompatible') pretraining is necessary at all by also comparing to the same architectures of the same size which get randomly initialised and then fine-tuned for the task. We denote those variants of each model as \textit{random}. 

\begin{figure}[t]
\centering
  \includegraphics[width=0.48\textwidth]{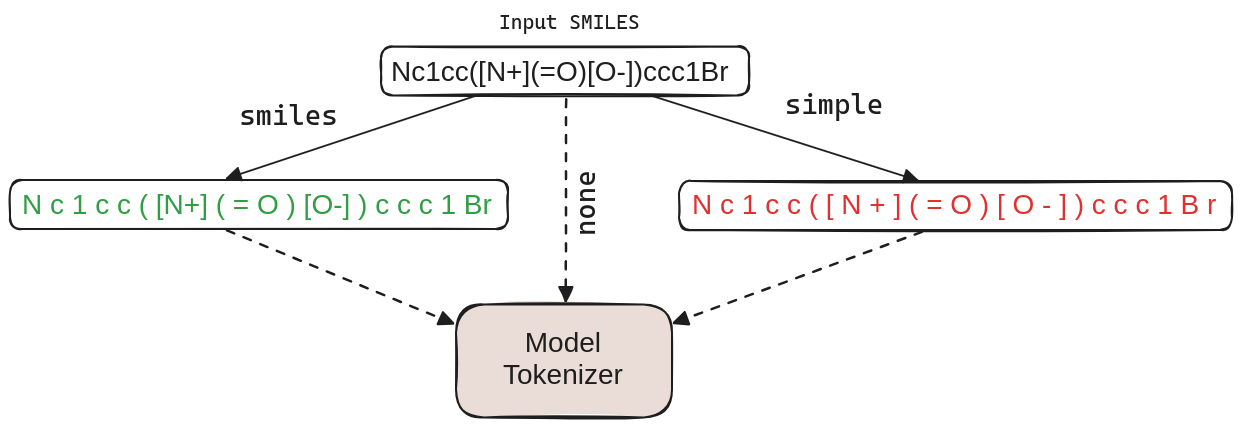}
  \caption{Illustration of different preprocessing strategies for SMILES input.}
  \label{fig:tokenisation}
\end{figure}

Furthermore, we also analyse whether continued pretraining in a self-supervised fashion on SMILES data offers any perfomance benefits before task-specific fine-tuning (denoted as \textit{cont}). For the continued pretraining we rely on the standard masked language modelling objective adapted to SMILES. For each SMILES we sample 15\% of its constituent tokens (where constituent tokens are based on the standard regular expression for SMILES from prior work \cite{Lu:2022unified}, see also later) and then we do one of the following options: (a) replace the token with a special mask token (the \textit{`\$'} character is used for the mask token) with the probability of 80\%; or (b) replace the token with another random token from the same SMILES (10\% probability); or (c) keep the token `as is' (10\% probability). 

Finally, we test whether T5-style models of the same sizes pretrained in a multi-task fashion to handle both text and SMILES generation simultaneously offer better initialisation points for single task-specific fine-tuning. For the latter, we select molT5 \cite{molt5} and nach0 \cite{nach0} (their \textit{Base} variants spanning 220M parameters) as two representative recent models from this `SMILES-aware' family of encoder-decoder models.

\rparagraph{Input Preprocessing, Vocabulary, and Tokenisation}
An often overlooked aspect of language models which may also have profound impact on model behaviour and final performance is the choice (and size) of vocabulary and corresponding tokenisation.\cite{tokeniser1,tokeniser2} We thus also delve deeper into their impact on final performance in organic reaction prediction tasks. With T5 and FlanT5 we experiment with different input preprocessing strategies, as also illustrated in Figure~\ref{fig:tokenisation}. First, given a SMILES sequence as input, we can decide to directly feed the sequence into the model tokeniser: we refer to this variant as \textit{+none}. We can also insert whitespaces into the input sequence in a trivial way, by simply dividing each character by a whitespace, ignoring any specific/chemistry domain knowledge; see Figure~\ref{fig:tokenisation}. This means that the subsequence \textit{Br} will be preprocessed into \textit{B r}, and e.g. \textit{[N+]} will be turned into \textit{[ N + ]}. We refer to this input preprocessing strategy as \textit{+simple}. It is possible to improve this via `SMILES-aware' preprocessing which relies on a set of standard regular expressions as used in prior work \cite{Lu:2022unified} to chunk the input sequence into a whitespace-separated set of tokens. This strategy, referred to as \textit{+smiles}, will not insert whitespaces into valid subsequences such as \textit{Br} and \textit{[N+]}. The preprocessed input is then fed to the standard tokeniser of the pretrained model which has been created on and for the natural language data. In this work, we thus test whether T5 and FlanT5 models can be used to process SMILES even with `ready-made' tokenisers without any adaptation or creation based on the (large collections of) raw SMILES data.

Further, the original vocabularies of FlanT5 (and T5) typically span $~32,000$ subwords, while most of these subwords are associated with natural language subwords and can be safely discarded when processing SMILES with a much more restricted vocabulary. We thus \textit{trim} the original vocabulary of FlanT5 by tokenising the large dataset of 116M SMILES sequences preprocessed via the \textit{+smiles} strategy above and then retaining only the subwords (and the corresponding embeddings) that occurred in those tokenised SMILES sequences.\footnote{In practice, we did not tokenise the entire set of 116M SMILES sequences but stopped tokenisation when we encountered that the set of `seen subwords' has not been extended after seeing 500k new SMILES sequences.} By doing this, we trimmed the vocabulary from the original 32k subwords to only 324 subwords, plus the three standard special tokens denoting the padding token, the start-of-sequence token and the `unknown' token. This trimming of the vocabulary and the corresponding embeddings has a double effect: \textbf{1)} it speeds up training and inference as it effectively constrains the search space, and \textbf{2)} it reduces model size from 60M to 44.4M parameters (FlanT5$_{Small}$) or from 220M to 198.7M (FlanT5$_{Base}$) without losing any modeling capability and expressiveness in organic reaction prediction tasks. We denote variants that rely on the vocabulary and embedding trimming step as \textit{+trim}, and variants with the original vocabulary as \textit{+orig}. Different model variants are then possible when this step is combined with input preprocessing strategies (\textit{+none+orig} as the simplest variant without any interventions versus, e.g., \textit{+smi+orig} or \textit{+smi+trim}).

Finally, given that SMILES comprise only special symbols, numbers, and alphabet letter where common atoms involved in organic reactions are represented as a single or double letters, namely \textit{Br, Cl, N, O, S, P, F, I, b, c, n, o, s, p} (Note: Upper case letters refer to non-aromatic atoms and lower case letters refer to aromatic atoms in SMILES), all the characters in SMILES sequences are UTF-8 compatible and can be easily fed to the tokenisation-free byte-level model such as ByT5. ByT5 is `tokenisation-free' as it does not rely on a learned vocabulary to map words or sub-word units to vocabulary items and simply operates on the vocabulary of 256 UTF-8 characters. Therefore, in this work we also analyse the potential of such byte-level natural language-pretrained encoder-decoder models for organic reaction prediction tasks. We run ByT5 without any input preprocessing to keep the length of the sequences tractable, e.g., the only variant tested is \textit{+none+orig}. In prior work in NLP, it was shown that ByT5 is significantly more robust to noise in the data and performs better on tasks that are sensitive to spelling, grammar errors and ambiguous expressions, such as text on social media platforms or speech transcribed to text.\cite{xtremeup} Our goal is to assess whether ByT5’s advantage related to dealing with noise in NLP data can be translated to better handling of noise in the chemical reaction data such as typically encountared in the USPTO datasets.


\rparagraph{Decoding Strategy}
In NLP, the use of the decoding strategy (i.e., (i.e., the
method used to generate strings from the model) )can have profound impact on output text~\cite{wiher-etal-2022-decoding}. However, the impact of the decoding strategy when working with SMILES as input has not been properly investigated: prior work \cite{Lu:2022unified} typically fixes the decoding strategy to the standard beam search (with beam width fixed, typically to 5), and does not provide any further evidence on how the chosen decoding strategy might impact the final output. In this paper, we thus compare the performance with much more efficient \textit{greedy search} decoding to the standard beam search, and also analyse how varying beam width impacts the final results. Moreover, we also run additional experiments with more sophisticated text generation strategies adopted from NLP research such as \textit{nucleus sampling}~\cite{nucleus} and \textit{contrastive search}~\cite{contrastive}. For technical details on the respective generation strategies, we refer the reader to the original publications.

\section{Experimental Setup}
\label{sec:expsetup}
We examine a range of encoder-decoder models with \textit{distinct properties} (e.g., tokenisation, model size, model architecture) in a \textit{variety of setups} (e.g., full-model fine-tuning, parameter-efficient fine-tuning, training data size, inference strategies). Further combined with a r\textit{ange of possible reaction prediction tasks} (e.g., forward reaction prediction, single-step retrosynthesis, reaction yield prediction, reaction type classification), this yields a huge space of possible experimental configurations. Therefore, due to large computational demands associated with the full experimental space, we do not present the full spectrum of results across all possible tasks; we rather zoom in and provide a representative selection of models, tasks, and experimental configurations that offer useful insight into the main interactions and models' inner workings without loss of generality.

\rparagraph{Target Tasks and Datasets}
Following prior work~\cite{Schwaller:2019molecular,Lu:2022unified}, we train and evaluate the models on:

\noindent \textbf{1) Forward reaction prediction (FWD-S)} with reactants-reagents separation \footnote{Preliminary experiments in the task version with reactants and reagents mixed\cite{Lu:2022unified} yields very similar relative trends in results and comparisons; we thus omit it for brevity and to save computation.} on the USPTO\_MIT dataset. The full training set consists of 409,035 input-output pairs. 

\noindent \textbf{2) Single-step retrosynthesis (RETRO)} on USPTO\_50k, where the full training set comprises 40,029 input-output pairs.

\noindent \textbf{3) Reagent prediction (REAG)} on the USPTO\_500\_MT dataset. While USPTO\_500\_MT has been created primarily as a multi-task dataset \cite{Lu:2022unified}, unless stated otherwise we conduct task-specific tuning for the single task using only its corresponding data, which comprises 116,360 input-output pairs.
We use the datasets and corresponding splits as provided by Lu et al~\cite{Lu:2022unified}. We evaluate all the models on the full test set of the RETRO task (5,004 pairs), while we randomly sample 10,000 test instances from the full, larger test sets of FWD-S and REAG to speed up inference due to a large number of experiments.\footnote{We have empirically validated that the relative trends in results do not change due to the test set sampling. We also run a smaller selection of models on full test sets to enable direct comparison to prior methods that operated on the same datasets.}

\rparagraph{Input Preprocessing}
We experiment with different input preprocessing strategies primarily for FlanT5 (with some experiments on T5) as illustrated in Figure~\ref{fig:tokenisation}. The maximum input sequence length is task-dependent and is defined based on the longest input sequence in each task-specific training set. Accordingly, the maximum output sequence length is also set according to the longest output sequence in the corresponding training sets. 

\rparagraph{Evaluation Metrics}
We report the standard \textit{accuracy at rank K (Acc@K)} scores, measuring if the gold sequence can be found in the top $K$ output sequences generated by the model. We select $K$-s following prior work (i.e., it is $\{1,3,5\}$ for RETRO and REAG, and $\{1,2,5\}$ for FWD-S).


\rparagraph{Continued Pretraining Setup}
For the \textit{cont} variants that run self-supervised continued pretraining on SMILES data, we randomly sample a pretraining set of 10M SMILES from the full set of $~116$M SMILES, and run continued pretraining for 400,000 steps. It might be possible that longer pretraining and with a larger pretraining set, as conducted e.g. by T5Chem \cite{Lu:2022unified} might yield to models better adapted to SMILES input and output, but that setup exceeds our computational resources. We run continued pretraining for the FlanT5${_{Base}}$\textit{+trim+smi} variant and ByT5${_{Base}}$; the batch size is set to 64 for both models, learning rate is seto to 0.0001 with inverse square root decay, and the optimiser used is Adafactor~\cite{adafactor}.

\rparagraph{Training Setup}
Due to a large number of experiments, we run a very constrained hyperparameter search, based on the development sets of FWD-S and RETRO and FlanT5$_{Small}$\textit{+none+orig} as the underlying model variant. The found hyper-parameters are then ported to all the other models: we acknowledge that finer-grained hyper-parameter optimisation is warranted as part of future research. Unless stated otherwise, we fine-tune with the learning rate set to $0.003$, batch size is $64$, weight decay is set to $0.01$, and warmup is set to $5,000$ steps. For FWD-S in the standard setting we fine-tune for $100,000$ steps, while we fine-tune for 100 epochs for RETRO (62.5K steps), and 50 epochs for REAG ($\sim$90K steps). The optimiser is Adafactor in all the experiments. For each model variant and task, we select the checkpoint based on performance on the corresponding development set: in most cases, it is the end checkpoint. 


\rparagraph{Inference Setup} The main decoding strategy is standard beam search with beam width set to 5, again following prior work \cite{Schwaller:2019molecular,Lu:2022unified,nach0}. In \S\ref{ss:further}, we also analyse the impact of beam width on Acc@1 scores, and also run preliminary investigations with more sophisticated decoding strategies borrowed from NLP research (see \S\ref{sec:methodology}).


\section{Results and Discussion}
\label{sec:results}
\input{tables/3tasks-orig-spaces}
\input{tables/flant5-fwd-configs}

\rparagraph{Impact of the Underlying Encoder-Decoder Model}
The results in the three tasks with different underlying encoder-decoder models are summarised in Table~\ref{tab:3tasks}. Several findings emerge from the reported scores. First, results from all the models are `in the same ballpark' although some variation across different models does exist. For instance, FlanT5 is a better initialisation point for task-specific fine-tuning than the corresponding original T5 model, yielding higher scores across all Acc@K metrics and all three tasks. Second, byte-level models offer the highest and most robust performance across the three tasks on average, indicating that byte-level `tokenisation-free' approach is a promising avenue for future work dealing with SMILES input and output. Next, the two models that were already fine-tuned on SMILES data and relevant tasks do not exhibit any advantage as starting points for task-specific fine-tuning over the `language-tuned' models such as FlanT5 and ByT5. The exception is the REAG task where molT5$_{Base}$ and nach0$_{Base}$ show strong performance: we suspect that this is due to the fact that the exact REAG training data was already included into their multi-task tuning, while they did not see the data of the other two tasks beforehand. Finally, a general important finding is that `language-pretrained' encoder-decoder models can be directly used for task-specific fine-tuning, even without any SMILES-oriented self-supervised pretraining, yielding competitive and robust performance.

Running the best-performing models on the full test set of FWD-S yields the peak Acc@1/Acc@5 scores of 90.2/96.2 which surpasses performance of Molecular Transformer\cite{Schwaller:2019molecular} (88.8/92.6) and is close to the T5-Chem model (90.4/96.4) despite the fact that the models have not been pretrained on SMILES prior to task-specific fine-tuning. The advantage over Molecular Transformer also remains on the full test set of RETRO (Acc@1 of 45.0 vs 43.5), but the gap to T5-Chem is slightly larger (45.0/66.7 vs 46.5/70.5).

\rparagraph{Impact of Input Preprocessing, Vocabulary Size and Tokenisation}
The results of the analysis, based on the FlanT5 model family, are provided in Table~\ref{tab:flant5}. We observe that trimming the vocabulary and the corresponding embeddings (i.e., the \textit{+trim+smi} variant) does not have any negative impact on the final performance: on the contrary, due to its side-effect of constraining the search space, it even has a slight positive impact on the final scores. In terms of input preprocessing, \textit{+smi} has a slight edge over \textit{+simple} and \textit{+none} (but the differences are sometimes minimal). This holds for FlanT5 of both tested sizes as well as for the original T5 model of those sizes (not shown). We speculate that this might be due to the fact that the language-pretrained model is better adapted to seeing input that contains whitespaces, but further investigations are warranted in future research. The \textit{random} variant without any language-specific pretraining is unable to learn the task well, even with relatively large amounts of training data (e.g., $~400K$ training instances for FWD-S), which again indicates the importance of non-random initialisation from the language-pretrained model. 

Finally, as expected, the \textit{+trim+none} variant provides extremely low scores: we report it for didactic purposes to emphasise how the mismatch between input preprocessing, vocabulary and tokenisation can have extremely detrimental negative impact: that variant trims the vocabulary based on the \textit{+spaces} preprocessing, while the actual input to the model does not undergo the same preprocessing step, which creates the mismatch and the model has to deal with (sub)sequences that result in the `unknown' tokens.

\rparagraph{Impact of Model Size}
Zooming again into Tables~\ref{tab:3tasks} and \ref{tab:flant5}, we observe that, as expected, the \textit{Base} variants of different models offer slightly higher performance than the \textit{Small} variants. This holds for T5 and FlanT5 as well as for ByT5 and molT5. However, the gap between \textit{Small} and \textit{Base} is not large (the only exception is T5 on REAG, see Table~\ref{tab:3tasks}): therefore, the choice of the model size also depends on the final need - if performance is paramount, \textit{Base} is a stronger option, while the \textit{Small} variants trade off some performance for higher training and inference efficiency.

In order to verify if further performance benefits can be reaped from an even larger encoder-decoder model, we fine-tune FlanT5$_{Large}$ (780M parameters) in the RETRO task. However, the increase in model size does not result in any increase in performance, with obtained Acc@1 / Acc@3 / Acc@5 scores of 43.96 / 59.61 / 64.23. In fact, the scores even decrease a bit, which might be the result of overfitting to the RETRO training data.

\rparagraph{Impact of Continued SMILES-Based Pretraining}
Continued pretraining (the \textit{+cont} variant in Table~\ref{tab:flant5}) does not have positive impact on final task performance in FWD-S and RETRO: in fact, it even yields slight performance drops. However, the \textit{+cont} variant of ByT5$_{Base}$ does yield some small gains in both tasks when we decrease the learning rate for fine-tuning from the default 0.003 to 0.0003: it reaches 90.37 / 93.9 (Acc@2) / 96.07 in FWD-S and 44.88 / 61.33 (Acc@3) / 66.29. These mixed preliminary results call for further investigation and also indicate the importance of finer-grained hyper-parameter optimisation, which is required as part of future research.

\rparagraph{Visualisation of Prediction with SHAP}
To check the chemistry validity of our models, we applied SHAP (SHapley Additive exPlanations)\cite{shap} to explain the predictions made by ByT5 and FlanT5 (\textit{+orig+none} variant of FlanT5 is used for simplicity). The SHAP explanations visualise the contributions of reactant and reagents to the structure of the predicted product at the token level. SHAP is a popular approach to explain the output of any machine learning model using game theory. The key idea behind SHAP is to calculate the Shapley values for each feature of the dataset and each Shapley value represents the impact of that feature on the model’s prediction. In the case of sequence input and output, such as the reactions we work with, multiple Shapley values are calculated for every token in the input sequence (i.e. reactant and reagents) for every token in the output sequence (i.e. the predict product). These SHAP values can be visualised in a matrix-like heatmap. 

\begin{figure}
    \centering
    \includegraphics[width=1\linewidth]{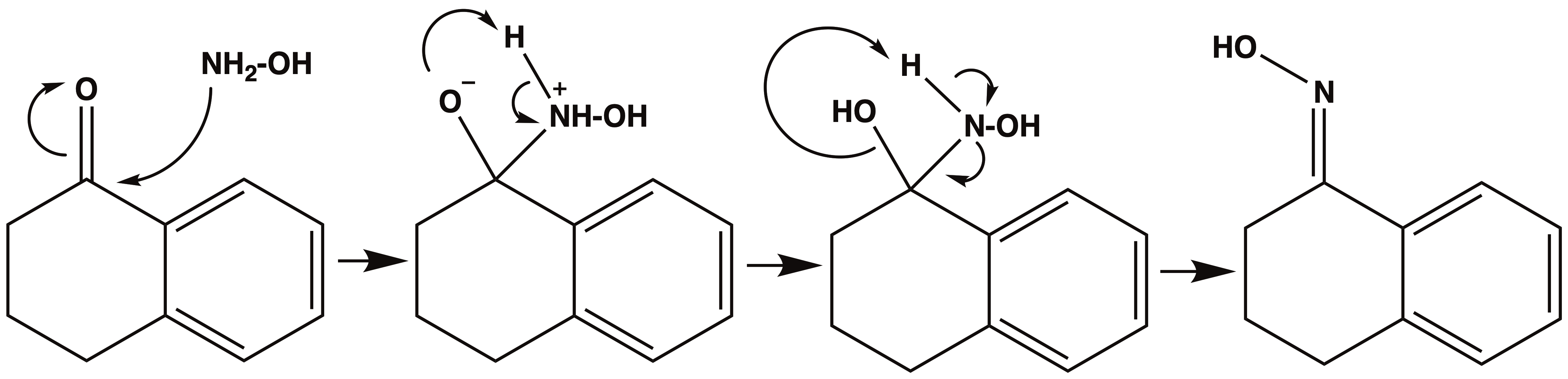}
    \caption{The reaction mechanism to generate ketoxime from ketone and hydroxylamine.}
    \label{fig:mechanism}
\end{figure}

We analysed a classic organic reaction: the generation of ketoxime from hydroxylamine and ketone in the presence of HCl and CH\textsubscript{3}OH. The SMILES representation of the reaction is as follows:
NO.O=C1CCCc2ccccc21.CO.Cl>>ON=C1CCCc2ccccc21
This reaction involves first the nucleophilic attack of the nitrogen in hydroxylamine (NH\textsubscript{2}-OH) on the carbonyl carbon in the ketone, followed by two successive proton transfers from the nitrogen to the oxygen in C=O to allow for elimination of water, resulting in the formation of the oxime functional group (C=N-OH) (Figure~\ref{fig:mechanism}). 

\begin{figure}[!ht]
    \centering
    \includegraphics[width=0.75\linewidth]{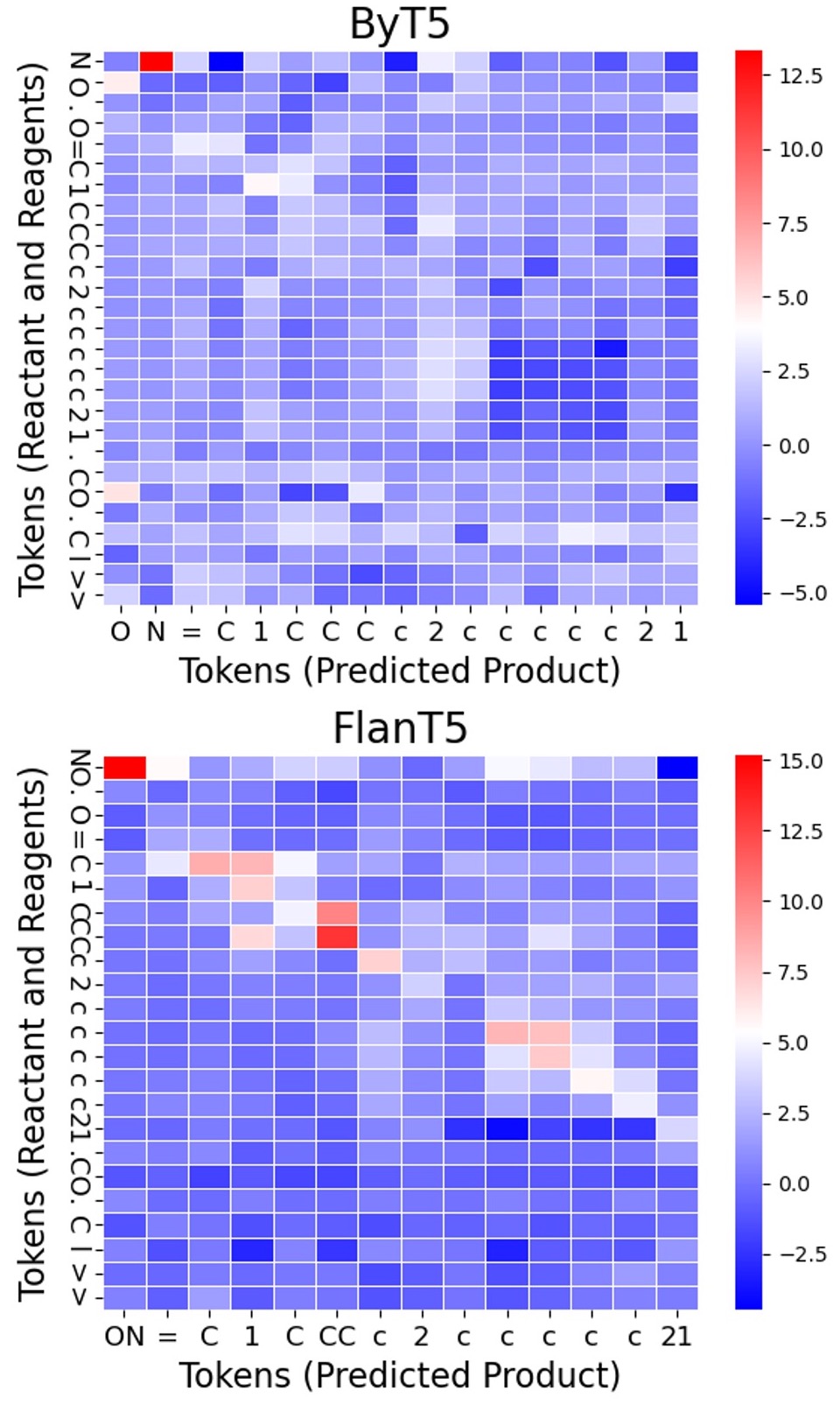}
    \caption{Computed Shapley values for the reaction }
    \label{fig:matrix}
\end{figure}

\begin{figure}[!th]
    \centering
    \includegraphics[width=1\linewidth]{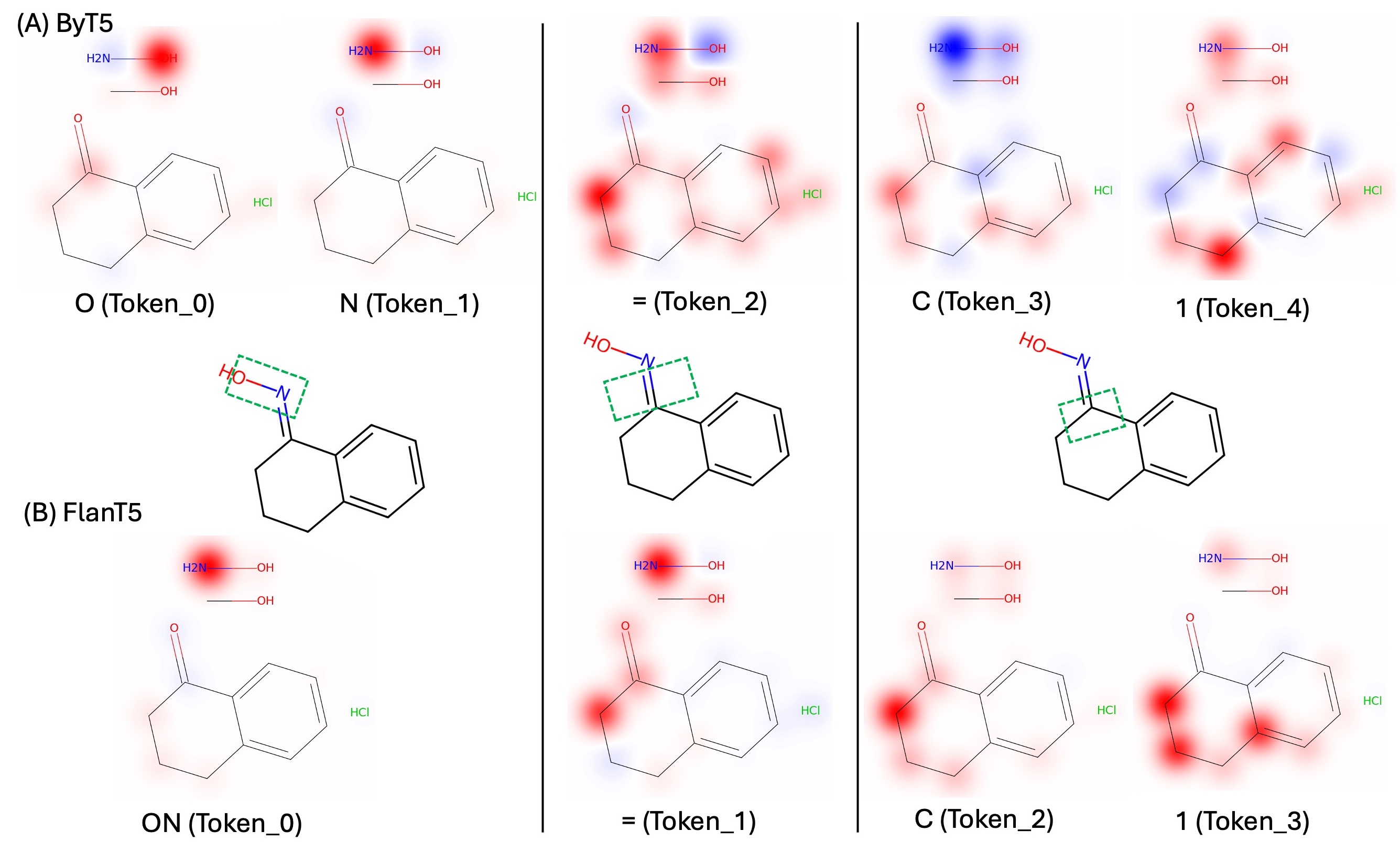}
    \caption{Visualisation of the impact of tokens in the reactants and reagents on the first few tokens in the predicted product}
    \label{fig:rdkit}
\end{figure}

Figure~\ref{fig:matrix} displays the computed Shapley values for this reaction from the ByT5 and FlanT5 models. The figure is generated using Seaborn heatmap and coloured using the ‘bwr’ colormap from matplotlib. The y-axis represents the sequence of tokens from the reactants and reagents, while the x-axis represents the sequence of tokens from the predicted product. Additionally, we visualised the impact of tokens in the reactants and reagents on the first few tokens in the product by projecting the Shapley values onto their 2d structures (Figure~\ref{fig:rdkit}). Figure~\ref{fig:rdkit} was generated using the GetSimilarityMapFromWeights function in RDkit\cite{rdkit} and coloured using the ‘bwr’ colormap from matplotlib. 

Both models highlight the hydroxylamine (N and O, token\_0 and token\_1 from ByT5; NO, token\_0 in FlanT5) as having the most significant impact on the product, which aligns with the underlying reaction mechanism. Furthermore, both models correctly identified that =N-OH in the product originates from the hydroxylamine, while the oxygen in the reactant ketone has a much weaker impact. This suggests that the models may have learned the correct reaction pathway, i.e. the oxygen in the ketone as the leaving group. In addition, atoms present in both the reactant and product (such as the aromatic ring and the cyclic aliphatic ketone) also exhibit noticeable impact. For the double bond in C1=N-OH, neighbouring carbons in the cyclic aliphatic ring have a strong impact. Although the exact reason for this is not clear, it is well known that S\textsubscript{N}2 reactions, like this one, are sensitive to steric hindrance from neighbouring atoms. It is likely that our models have learned correlations between key substructures of the molecule, as has been observed in T5Chem. 

\subsection{Further Analyses}
\label{ss:further}
%
%
\sparagraph{Data Efficiency Experiments}
We run data efficiency experiments on the FWD-S task, where we subsample smaller training data, taking $1/2$ ($204,518$ training instances), $1/4, \ldots,$$ 1/128$ ($3,196$ training instances) of the full FWD-S training set. Each smaller set is a subset of all the larger (sub)sets. The models in comparison are \textit{Base} versions of FlanT5 (\textit{+trim+smi}), ByT5, and molT5: the summary of their performance across different FWD-S data sizes is provided in Figure~\ref{fig:sample_eff}. The three models display very similar `performance trajectories' where molT5 lags slightly behind the two other models for smaller data samples across both Acc@K metrics. Combined with the results already observed in Tables~\ref{tab:3tasks} and \ref{tab:flant5}, these plots point to a more general conclusion that the actual training data size and quality is more instrumental to the final performance than the chosen encoder-decoder architecture. While there is some variation that stems from the choice of input preprocessing, tokenisation, and model size, the data size is still a key factor determining the `performance magnitude'. This conjecture calls for more future research across two axes: \textbf{1)} creation of larger and higher-quality training and evaluation datasets~\cite{nach0} for single-task and multi-task training; \textbf{2)} work on more sample-efficient learning via transfer learning~\cite{}. 

\input{figures/sample_eff}


\rparagraph{Impact of the Decoding Strategy} We vary beam width from 1 (greedy search) to 10 for two high-performing model variants in FWD-S and RETRO and plot Acc@1 and Acc@5 scores in Figure~\ref{fig:beam_analysis}. Very similar patterns have been observed also for other model variants across different tasks. First, we note that the most efficient decoding strategy, greedy search, is already very competitive in terms of Acc@1 scores and only marginal gains can be achieved with beam search. Moreover, Acc@1 with larger beam sizes saturates quickly and the peak Acc@1 score is typically achieved already with beam size set to 2 or 3. This basically indicates that for efficiency reasons, if Acc@1 is paramount, there is typically no need to increase the beam size. Moreover, for larger beam sizes ($>5$), Acc@5 scores even seem to slightly decrease as the model might provide more exploratory generations. 

Additional experiments with top-K sampling, nucleus sampling and contrastive search with hyper-parameters suggested from NLP research (e.g., temperature for top-K sampling, top\_p for nucleus sampling) did not yield any noteworthy benefits over the simple greedy search. Changing the hyper-parameters for the efficient nucleus sampling based on development set performance can yield slight benefits at inference over greedy search without damaging inference efficiency, but the gains are typically slight, ranging between 0.1 and 0.4 Acc@1 points. In sum, a more focused study on the impact of decoding strategy for SMILES generation is also warranted as part of future research.

\input{figures/beam_analysis}

\input{tables/gpu-timing}
\rparagraph{Estimate of Training Time on GPUs}
We provide rough estimate of \textit{wall-clock GPU time} for several model variants in the training-wise most demanding FWD-S task (400,000+ training instances) in Table~\ref{tab:gpu_time}. The estimates have been made in the following setup for all models: 100,000 training steps with the batch size of 16 and gradient accumulation of 4 (yielding the effective batch size of 64), input sequence length and output sequence length set to 144. All the estimates are based on single runs on a single 24GiB NVIDIA RTX 4090 GPU. The time estimates indicate that the main experiments with the chosen model architectures and their corresponding size can typically be run on consumer-level GPUs. Another finding is that some speed-ups can be achieved via trimming the vocabulary and the corresponding embeddings (moving from \textit{+orig} to \textit{+trim}) without any performance degradation (see Table~\ref{tab:flant5} again). Concerning inference time, our previous experiments with decoding strategies indicate that, especially if top-1 accuracy is paramount, a better trade-off between inference efficiency and performance can be struck with smaller beam width or even with simple greedy search.

\rparagraph{On Multi-Task Fine-Tuning}
While all previous experiments focused on task-specific single-task fine-tuning which yields models specialized for a single task, we also briefly test whether pretrained language models such as ByT5 and FlanT5 can be readily used for multi-task fine-tuning as well, even without any modification of the fine-tuning protocol. To this end, we rely on exactly the same setup and hyper-parameters as with single-task tuning previously (see \S\ref{sec:expsetup}) and we run multi-task fine-tuning for 100,000 training steps on multi-task data of the USPTO\_500\_MT dataset~\cite{Lu:2022unified}: it again covers three tasks (forward reaction prediction, retrosynthesis, and reagent prediction), where training data of each task constitutes of 116,360 instances for the total of 349,080 training instances for multi-task training. Each instance is marked by a specific prefix (\textit{Product:}, \textit{Reactants:}, \textit{Reagents:}) which links it to the origin task. 

We fine-tune two models: FlanT5$_{Base}$ (\textit{+orig+none} variant for simplicity) and ByT5$_{Small}$, and run standard evaluation on the three tasks. Acc@1 scores of the multi-task model on the three tasks are 24.01 (REAG), 95.82 (FWD), 71.77 for FlanT5$_{Base}$, and 24.15, 96.78, 72.24 for ByT5$_{Small}$. This indicates that multi-task fine-tuning is also possible starting from language-pretrained model checkpoints. We then continue to fine-tune the ByT5$_{Small}$ multi-task for the single REAG task. This yields marginal gains in the REAG task (from 24.15 to 24.38 with greedy search), but, as expected, it also yields catastrophic forgetting of the the other two tasks (e.g., Acc@1 drops from 95.82  to 35.26) due to full-model specialisation.

Motivated by these preliminary results, we plan to delve deeper into multi-task fine-tuning setups in future work, also coupled with recent advances in modular and parameter-efficient learning in NLP that by design avoid issues such as catastrophic forgetting and interfence in multi-task setups~\cite{peft,delta,modular}.

\section{Reflection and Conclusion}
\label{sec:conclusion}
Our work operates at the intersection of several broad areas of NLP research and AI in chemistry, including multi-task learning, transfer learning and computer-assisted synthesis planning. Our preliminary results indicate that although FlanT5 and ByT5 are pretrained only on language tasks, they provide a solid foundation for fine-tuning in reaction prediction. This suggests that GPU-extensive pretraining on large unlabelled molecules may not be essential to leverage the power of these state-of-the-art language models for chemistry. While there is some variation that stems from the choice of input preprocessing, tokenisation, and model size, the training data size is instrumental to the final performance of the fine-tuned models. It is worth noting that the USPTO\_500\_MT dataset used in our work (a subset curated from the full USPTO (1976-2013) dataset\cite{Lowe2017}) covers only general organic reactions. It contains limited stereochemical information\cite{Pesciullesi2020} and data on catalysts is scarce. Further fine-tuning will be necessary to specialise our models to specific types of reactions or prediction tasks. In such cases, factors that are not significant to the performance of our models could be explored further. For example, when predicting catalysts, a larger beam size and perhaps more sophisticated decoding strategies will be needed to provide more exploratory predicted output. For future work, we plan to delve deeper into multi-task fine-tuning using recent advances in parameter-efficient and modular learning and improve our models for novel and more challenging reaction prediction tasks. 

\rparagraph{Limitations}
One limitation of this wide empirical study is that, due to a large number of experiments coupled with computational constraints, all the experimental configurations were run once (with a random seed set to 42), while averaging over multiple training runs is a typical good practice~\cite{dodge-etal-2019-show}. However, in our preliminary experiments, we verified that score fluctuations were minimal, which should mitigate this concern. Furthermore, as mentioned in \S\ref{sec:expsetup}, a larger-scale per-model and per-variant hyper-parameter optimisation might also yield improved performance. Improved performance may also be achieved via longer fine-tuning and checkpoint selection using the development sets. We also have not extensively explored the full range of possible decoding strategies, nor have we thoroughly optimised hyper-parameters associated with the tested decoding strategies (e.g., for nucleus sampling): this can also be further tuned on the task-specific development sets. Future work should also further analyse the interplay between longer and more GPU-intensive self-supervised pretraining on larger SMILES data and task-specific fine-tuning.


\section*{Author Contributions}
JP and IV conceived and designed the study, carried out the research and wrote the manuscript.

\section*{Conflicts of interest}
There are no conflicts to declare.

\section*{Acknowledgements}
The work was supported by the UK Engineering and Physical Sciences Research Council (EPSRC) (grant number EP/Y004167/1). Ivan Vuli\'{c} would like to thank Benjamin Minixhofer for some fruitful discussions on tokenisation.

\bibliography{rsc} 
\bibliographystyle{rsc} 

\end{document}

%% file: _custom-things.tex
\usepackage{xspace}
\usepackage{enumitem}
\usepackage{booktabs}
\usepackage{arydshln}
\usepackage{tabularx}
\usepackage{colortbl}
\usepackage{subcaption}

\newcommand{\rparagraph}[1]{\vspace{1.2mm}\noindent\textbf{#1.}}

\newcommand{\sparagraph}[1]{\vspace{0.0mm}\noindent\textbf{#1.}}

\newcommand{\rparagraphnodot}[1]{\vspace{1.2mm}\noindent\textbf{#1}}

\definecolor{Gray}{gray}{0.92}
\definecolor{racing-green}{rgb}{0.0, 0.8, 0.6}
\definecolor{awesome-red}{rgb}{1.0, 0.13, 0.32}
\newcolumntype{Y}{>{\centering\arraybackslash}X}

%% file: _style-addons.tex
\pagestyle{fancy}
\thispagestyle{plain}
\fancypagestyle{plain}{
\renewcommand{\headrulewidth}{0pt}
}

\makeFNbottom
\makeatletter
\renewcommand\LARGE{\@setfontsize\LARGE{15pt}{17}}
\renewcommand\Large{\@setfontsize\Large{12pt}{14}}
\renewcommand\large{\@setfontsize\large{10pt}{12}}
\renewcommand\footnotesize{\@setfontsize\footnotesize{7pt}{10}}
\makeatother

\renewcommand{\thefootnote}{\fnsymbol{footnote}}
\renewcommand\footnoterule{\vspace*{1pt}%
\color{cream}\hrule width 3.5in height 0.4pt \color{black}\vspace*{5pt}} 
\setcounter{secnumdepth}{5}

\makeatletter 
\renewcommand\@biblabel[1]{#1}            
\renewcommand\@makefntext[1]%
{\noindent\makebox[0pt][r]{\@thefnmark\,}#1}
\makeatother 
\renewcommand{\figurename}{\small{Fig.}~}
\sectionfont{\sffamily\Large}
\subsectionfont{\normalsize}
\subsubsectionfont{\bf}
\setstretch{1.125} 
\setlength{\skip\footins}{0.8cm}
\setlength{\footnotesep}{0.25cm}
\setlength{\jot}{10pt}
\titlespacing*{\section}{0pt}{4pt}{4pt}
\titlespacing*{\subsection}{0pt}{15pt}{1pt}

\fancyfoot{}
\fancyfoot[LO,RE]{\vspace{-7.1pt}\includegraphics[height=9pt]{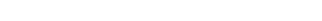}}
\fancyfoot[CO]{\vspace{-7.1pt}\hspace{13.2cm}\includegraphics{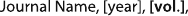}}
\fancyfoot[CE]{\vspace{-7.2pt}\hspace{-14.2cm}\includegraphics{head_foot/RF}}
\fancyfoot[RO]{\footnotesize{\sffamily{1--\pageref{LastPage} ~\textbar  \hspace{2pt}\thepage}}}
\fancyfoot[LE]{\footnotesize{\sffamily{\thepage~\textbar\hspace{3.45cm} 1--\pageref{LastPage}}}}
\fancyhead{}
\renewcommand{\headrulewidth}{0pt} 
\renewcommand{\footrulewidth}{0pt}
\setlength{\arrayrulewidth}{1pt}
\setlength{\columnsep}{6.5mm}
\setlength\bibsep{1pt}

\makeatletter 
\newlength{\figrulesep} 
\setlength{\figrulesep}{0.5\textfloatsep} 

\newcommand{\topfigrule}{\vspace*{-1pt}%
\noindent{\color{cream}\rule[-\figrulesep]{\columnwidth}{1.5pt}} }

\newcommand{\botfigrule}{\vspace*{-2pt}%
\noindent{\color{cream}\rule[\figrulesep]{\columnwidth}{1.5pt}} }

\newcommand{\dblfigrule}{\vspace*{-1pt}%
\noindent{\color{cream}\rule[-\figrulesep]{\textwidth}{1.5pt}} }

\makeatother

\twocolumn[
  \begin{@twocolumnfalse}
{\includegraphics[height=30pt]{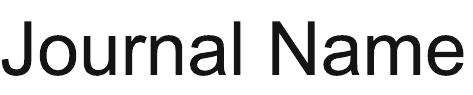}\hfill\raisebox{0pt}[0pt][0pt]{\includegraphics[height=55pt]{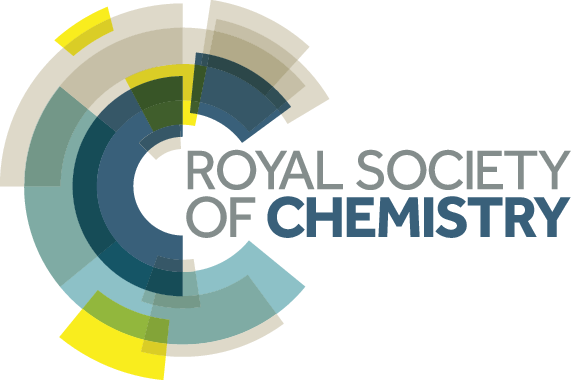}}\\[1ex]
\includegraphics[width=18.5cm]{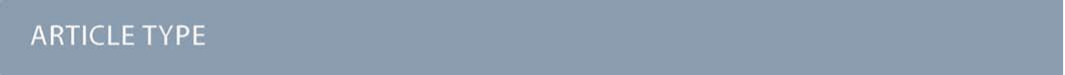}}\par
\vspace{1em}
\sffamily
\begin{tabular}{m{4.5cm} p{13.5cm} }

\includegraphics{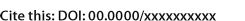} & \noindent\LARGE{\textbf{Specialising and Analysing Instruction-Tuned and Byte-Level Language Models for Organic Reaction Prediction}} \\
\vspace{0.3cm} & \vspace{0.3cm} \\

 & \noindent\large{Jiayun Pang,$^{\ast}$\textit{$^{a}$} and Ivan Vuli\'{c},$^{\ast}$\textit{$^{b}$}} \\

\includegraphics{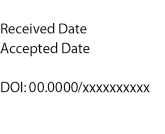} & \noindent\normalsize{

 \textit{}Transformer-based encoder-decoder models have demonstrated impressive results in chemical reaction prediction tasks. However, these models typically rely on pretraining using tens of millions of unlabelled molecules, which can be time-consuming and GPU-intensive. One of the central questions we aim to answer in this work is: Can FlanT5 and ByT5, the encode-decoder models pretrained solely on language data, be effectively specialised for organic reaction prediction through task-specific fine-tuning? We conduct a systematic empirical study on several key issues of the process, including tokenisation, the impact of (SMILES-oriented) pretraining, fine-tuning sample efficiency, and decoding algorithms at inference. Our key findings indicate that although being pretrained only on language tasks, FlanT5 and ByT5 provide a solid foundation to fine-tune for reaction prediction, and thus become `chemistry domain compatible' in the process. This suggests that GPU-intensive and expensive pretraining on a large dataset of unlabelled molecules may be useful yet not essential to leverage the power of language models for chemistry. All our models achieve comparable Top-1 and Top-5 accuracy although some variation across different models does exist. Notably, tokenisation and vocabulary trimming slightly affect final performance but can speed up training and inference; The most efficient greedy decoding strategy is very competitive while only marginal gains can be achieved from more sophisticated decoding algorithms. In summary, we evaluate FlanT5 and ByT5 across several dimensions and benchmark their impact on organic reaction prediction, which may guide more effective use of these state-of-the-art language models for chemistry-related tasks in the future. } \\

\end{tabular}

 \end{@twocolumnfalse} \vspace{0.6cm}

  ]

\renewcommand*\rmdefault{bch}\normalfont\upshape
\rmfamily
\section*{}
\vspace{-1cm}


\footnotetext{\textit{$^{a}$~School of Science, Faculty of Engineering and Science, University of Greenwich, Medway Campus, Central Avenue, Chatham Maritime, ME4 3RL, UK. E-mail: j.pang@gre.ac.uk}}
\footnotetext{\textit{$^{b}$~Language Technology Lab, University of Cambridge, 9 West Road, Cambridge CB3 9DA, UK. E-mail: iv250@cam.ac.uk }}




%% file: tables/3tasks-orig-spaces.tex
\begin{table*}[!th]
    \centering
    \caption{Results in the three evaluation tasks (see \S\ref{sec:expsetup}) with different encoder-decoder architectures (see \S\ref{sec:methodology}) trained on the full single-task training set. T5 and FlanT5 use the following configuration: original vocabularies (\textit{+orig}) with `SMILES spaces' (\textit{+smi}). All the results are obtained with standard beam search (beam size of 5). Peak scores per column are in boldface}
    \label{tab:3tasks}
    {\fontsize{9.6pt}{9.5pt}\selectfont
    \begin{tabularx}{\textwidth}{l YYY YYY YYY}
    \toprule
    {\bf Model$\downarrow$} & \multicolumn{3}{c}{FWD-S}  & \multicolumn{3}{c}{RETRO}  & \multicolumn{3}{c}{REAG} \\
    \cmidrule(lr){2-4} \cmidrule(lr){5-7} \cmidrule(lr){8-10}
    {} &  {Acc@1} & {Acc@2} & {Acc@5} & {Acc@1} & {Acc@3} & {Acc@5} & {Acc@1} & {Acc@3} & {Acc@5} \\
     \cmidrule(lr){2-4} \cmidrule(lr){5-7} \cmidrule(lr){8-10}
    {T5$_{Small}$} & {89.01} & {93.36} & {95.45} & {42.35} & {57.97} & {63.13} & {3.85} & {7.36} & {9.32} \\
    {T5$_{Base}$} & {89.28} & {93.36} & {95.47} & {42.59} & {58.31} & {62.89} & {20.33} & {29.72} & {33.90} \\
    \hdashline
    {FlanT5$_{Base}$} & {89.83} & {93.73} & {95.73} & {\bf 44.86} & {\bf 61.45} & {\bf 66.55} & {23.27} & {31.86} & {35.82} \\
    \hdashline
    {ByT5$_{Small}$} & {90.06} & {93.75} & {95.71} & {43.96} & {58.81} & {63.09} & {22.85} & {31.20} & {35.43} \\
    {ByT5$_{Base}$} & {\bf 90.10} & {\bf 93.90} & {\bf 96.07} & {44.74} & {60.25} & {64.89} & {24.18} & {32.27} & {36.18} \\
    \hdashline
    \hdashline
    {molT5$_{Small}$}\cite{molt5} & {88.98} & {93.23} & {95.60} & {42.63} & {59.09} & {63.53} & {20.89} & {27.81} & {31.39} \\
    {molT5$_{Base}$}\cite{molt5} & {89.90} & {93.68} & {95.75} & {42.71} & {58.45} & {63.77} & {25.0} & {32.87} & {36.82} \\
    {nach0$_{Base}$}\cite{nach0} & {87.33} & {92.12} & {94.72} & {41.33} & {57.35} & {62.59} & {\bf 25.0} & {\bf 33.54} & {\bf 37.26} \\
    \bottomrule
    \end{tabularx}
    }
\end{table*}

%% file: tables/flant5-fwd-configs.tex
\begin{table*}[!ht]
    \centering
    \caption{Analysis of different variants (varying vocabulary and tokenization, see \S\ref{sec:methodology}) of FlanT5 in the forward prediction and single-step retrosynthesis tasks. \textit{random} denotes a FlanT5 architecture of the same size and structure as the corresponding $_{Small}$ and $_{Base}$ model, but with randomly initialised parameters before the model undergoes task-specific FWD-S or RETRO fine-tuning. \textit{cont} denotes FlanT5 which was further fine-tuned in a self-supervised setup with the masked SMILES modeling objective (see \S\ref{sec:methodology}) before undergoing additional task-specific fine-tuning, and we run it only for FlanT5$_{Base}$.}
    \label{tab:flant5}
    {\fontsize{8.3pt}{8.2pt}\selectfont
    \begin{tabularx}{\textwidth}{l YYY YYY YYY YYY}
       \toprule
    {} & \multicolumn{6}{c}{FlanT5$_{Small}$} & \multicolumn{6}{c}{FlanT5$_{Base}$} \\
    \cmidrule(lr){2-7} \cmidrule(lr){8-13}
     {\bf Variant$\downarrow$} & \multicolumn{3}{c}{FWD-S} & \multicolumn{3}{c}{RETRO} & \multicolumn{3}{c}{FWD-S} & \multicolumn{3}{c}{RETRO} \\
    \cmidrule(lr){2-4} \cmidrule(lr){5-7} \cmidrule(lr){8-10} \cmidrule(lr){11-13}
    {} & {Acc@1} & {Acc@2} & {Acc@5} & {Acc@1} & {Acc@2} & {Acc@5} & {Acc@1} & {Acc@3} & {Acc@5} & {Acc@1} & {Acc@3} & {Acc@5}\\
    \cmidrule(lr){2-4} \cmidrule(lr){5-7} \cmidrule(lr){8-10} \cmidrule(lr){11-13}
    {\em +orig+smi} & {88.92} & {93.18} & {95.59} & {42.53} & {59.79} & {65.73} & {89.83} & {93.73} & {95.73} & {42.87} & {58.47} & {63.83} \\
    {\em +trim+smi} & {89.32} & {93.29} & {95.66} & {44.68} & {60.71} & {66.13} & {89.91} & {93.72} & {95.70} & {44.86} & {61.45} & {66.55} \\
    {\em +orig+none} & {88.70} & {93.22} & {95.56} & {43.07} & {58.05} & {63.21} & {89.84} & {93.51} & {95.58} & {42.03} & {57.71} & {62.05} \\
    {\em +trim+none} & {0.03} & {0.04} & {0.06} & {0.16} & {0.22} & {0.22} & {0.04} & {0.05} & {0.07} & {0.16} & {0.2} & {0.24} \\
    \hdashline
    {\em cont+trim+smi} & {--} & {--} & {--} & {--} & {--} & {--} & {89.34} & {93.39} & {95.48} & {43.55} & {59.89} & {65.17} \\
    \hdashline
    {\em +orig+simple} & {88.95} & {93.31} & {95.52} & {43.86} & {60.71} & {65.67} & {89.25} & {93.31} & {95.52} & {42.61} & {58.83} & {64.39}\\
    \hdashline
    {\em random} & {0.0} & {0.0} & {0.0} & {0.0} & {0.0} & {0.0} & {0.0} & {0.0} & {0.0} & {0.0} & {0.0} & {0.0} \\
    \bottomrule
    \end{tabularx}
    }
\end{table*}

%% file: figures/sample_eff.tex
\begin{figure}[t]
    \centering
    \begin{subfigure}[t]{.489\linewidth}
    \centering
        \includegraphics[width=.99\linewidth]{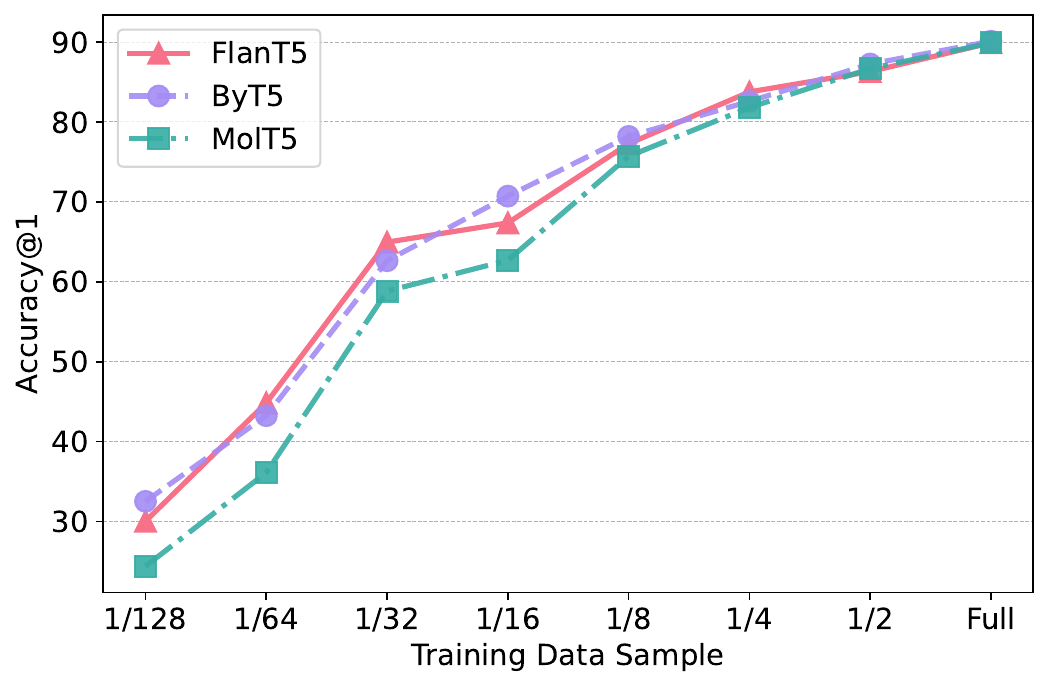}  
        \caption{Acc@1 Scores}
    \label{fig:CaseStudy:Polytropon}
    \end{subfigure}
    \begin{subfigure}[t]{.489\linewidth}
    \centering
        \includegraphics[width=.99\linewidth]{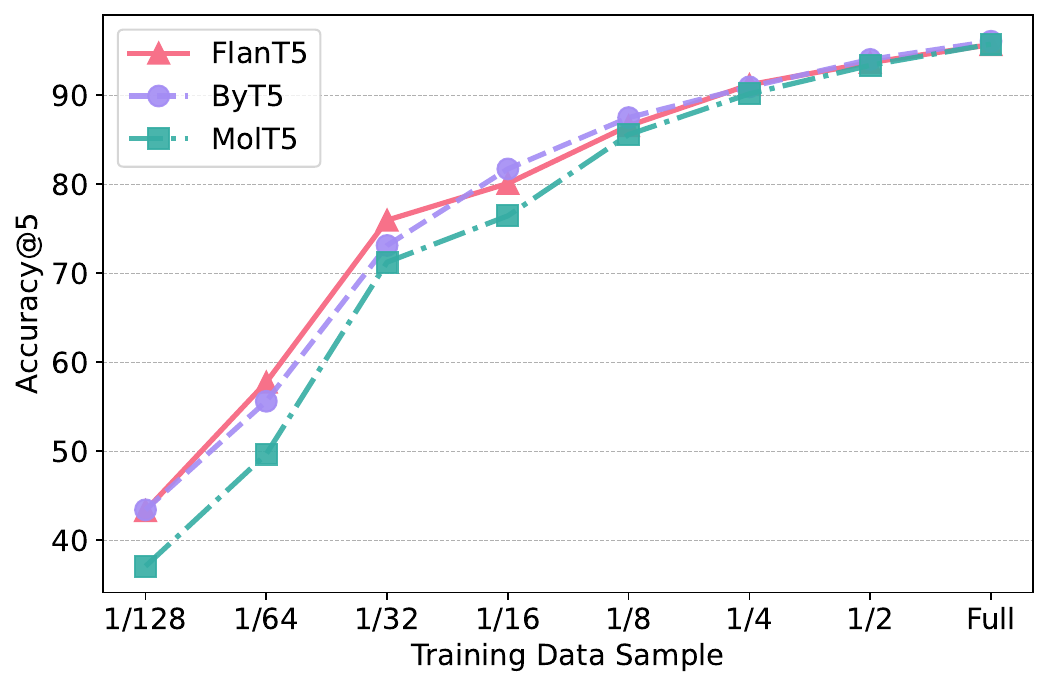}  
        \caption{Acc@5 Scores}
        \label{fig:CaseStudy:MoE}
    \end{subfigure}
    \caption{Data efficiency experiments; (a) Acc@1 and (b) Acc@5 scores in the FWD-S task on the same test set of 10,000 instances while varying the size of the training set (taking the fraction of the full set as denoted in the x axis); the FlanT5 variant is \textit{+trim+smi}
    }
\label{fig:sample_eff}
\end{figure}

%% file: figures/beam_analysis.tex
\begin{figure}[t]
    \centering
    \begin{subfigure}[t]{.489\linewidth}
    \centering
        \includegraphics[width=.99\linewidth]{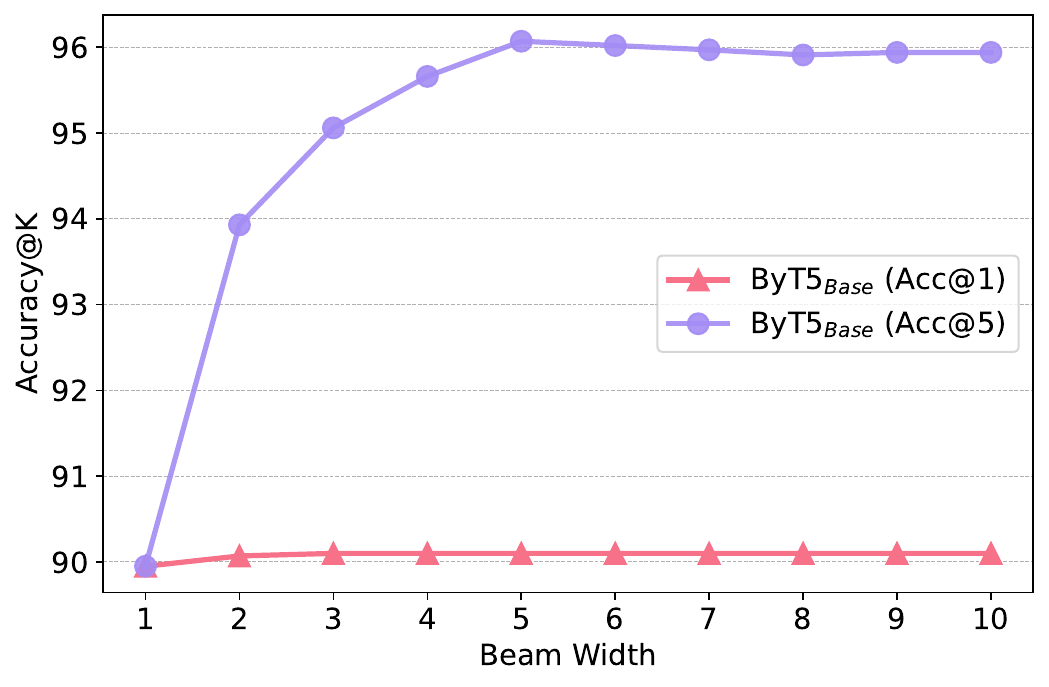}  
        \caption{FWD-S}
    \label{fig:beam_fwd}
    \end{subfigure}
    \begin{subfigure}[t]{.489\linewidth}
    \centering
        \includegraphics[width=.99\linewidth]{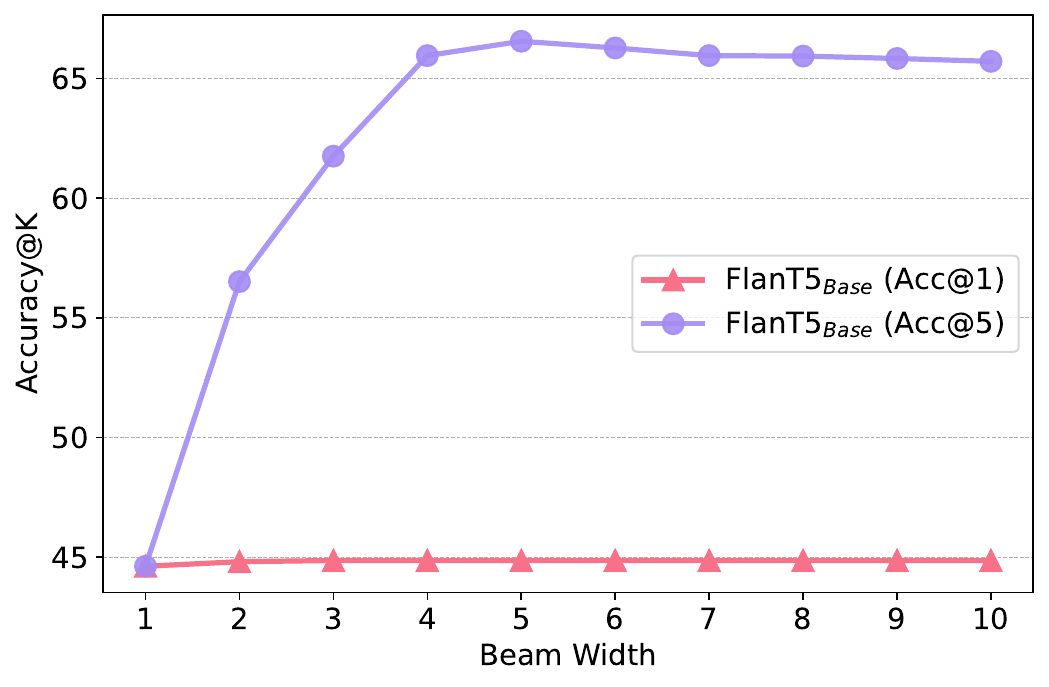}  
        \caption{RETRO}
        \label{fig:beam_retro}
    \end{subfigure}
    \caption{Beam analysis experiments; (a) FWD-S; (b) RETRO; we vary beam size from 1 (greedy) to 10; the FlanT5 variant is \textit{+trim+smi}
    }
\label{fig:beam_analysis}
\end{figure}

%% file: tables/gpu-timing.tex
\begin{table}[t!]
    \centering
    \caption{Wall-clock time estimates for the full training run (100,000 training steps) in the FWD-S task on a single 24GiB RTX 4090 GPU}
    \label{tab:gpu_time}
    {\fontsize{9.6pt}{9.5pt}\selectfont
    \begin{tabularx}{0.48\textwidth}{l Y}
    \toprule
    {\bf Model Variant} & {\bf Time} \\
    \cmidrule(lr){1-2}
    {FlanT5$_{Small}$+\textit{orig+smi}} & {5h 20m} \\
    {FlanT5$_{Small}$+\textit{trim+smi}} & {4h 25m} \\
    {FlanT5$_{Base}$+\textit{orig+smi}} & {14h 50m} \\
    {FlanT5$_{Base}$+\textit{trim+smi}} & {13h 35m} \\
    \hdashline
    {ByT5$_{Small}$} & {16h 20m} \\
    {ByT5$_{Base}$} & {30h 55m} \\
    \bottomrule
    \end{tabularx}
    }
\end{table}